\pdfoutput=1

\documentclass[11pt]{article}

\usepackage[]{ACL2023}

\usepackage{times}
\usepackage{latexsym}

\usepackage[T1]{fontenc}

\usepackage[utf8]{inputenc}
\usepackage{booktabs}
\usepackage{microtype}
\usepackage{times}  
\usepackage{helvet}  
\usepackage{graphicx}

\usepackage{inconsolata}
\usepackage{algorithm}
\usepackage{algorithmic}
\usepackage{amsmath}
\usepackage{amsfonts}
\usepackage{multirow}
\usepackage{multicol}
\usepackage{subfigure}
\usepackage{natbib}  
\usepackage{caption} 
\title{ECLM: Entity Level Language Model for Spoken Language 
\\ Understanding with Chain of Intent}

\author{
    \textbf{Shangjian Yin}$^{1}$,
    \textbf{Peijie Huang}$^{1}$\thanks{$~~$ Corresponding author.}, 
    \textbf{Jiatian Chen}$^{1}$ 
    \textbf{Haojing Huang}$^{2}$,
    \textbf{Yuhong Xu}$^{1}$, \\
    $^{1}$College of Mathematics and Informatics, South China Agricultural University, China \\
    $^{2}$Tsinghua Shenzhen International Graduate School, Tsinghua University \\
    \texttt{sjy8460@163.com, pjhuang@scau.edu.cn, c2541421012@163.com}, \\
    \texttt{hhj23@mails.tsinghua.edu.cn, xuyuhong@scau.edu.cn}
}

\begin{document}
\maketitle



\begin{abstract}
Large Language Models (LLMs) have demonstrated impressive capabilities in language generation and general task performance. However, their application to spoken language understanding (SLU) remains challenging, particularly for token-level tasks, where the autoregressive nature of LLMs often leads to misalignment issues. They also struggle to capture nuanced interrelations in semantic-level tasks through direct fine-tuning alone. To address these challenges, we propose the Entity-level Language Model (ECLM) framework, which reformulates slot-filling as an entity recognition task and introduces a novel concept, \textit{Chain of Intent}, to enable step-by-step multi-intent recognition. Experimental results show that ECLM significantly outperforms strong baselines such as Uni-MIS, achieving gains of 3.7\% on MixATIS and 3.1\% on MixSNIPS. Compared to standard supervised fine-tuning of LLMs, ECLM further achieves improvements of 8.5\% and 21.2\% on these datasets, respectively. Our code is available at \url{https://github.com/SJY8460/ECLM}.
\end{abstract}





\section{Introduction}

The rapid advancement of large language models (LLMs) has markedly accelerated progress in the field of natural language processing (NLP) \cite{Geogle,llama2}. Trained on extensive datasets, these models demonstrate exceptional performance across a wide range of NLP tasks, including natural language inference, summarization, and dialog systems, often achieving impressive results through in-context learning alone \cite{DBLP:conf/emnlp/Hu0X0SO22, DBLP:conf/eacl/KavumbaBHI23}.

Spoken language understanding (SLU) is a critical component of task-oriented dialog systems, which are designed to construct a semantic frame that accurately captures the user's request. This semantic frame is typically built through two sub-tasks: intent detection, which identifies the user's intent, and slot filling, which extracts relevant semantic elements. Given the close interdependence of these sub-tasks \citep{Tur2011SpokenLU}, state-of-the-art SLU systems often employ joint models to effectively capture the correlations between them \citep{DBLP:conf/naacl/GooGHHCHC18,DBLP:conf/emnlp/QinCLWL19}.

\begin{figure}[t] 
    \centering  
    \centerline{\includegraphics[width=1\linewidth]{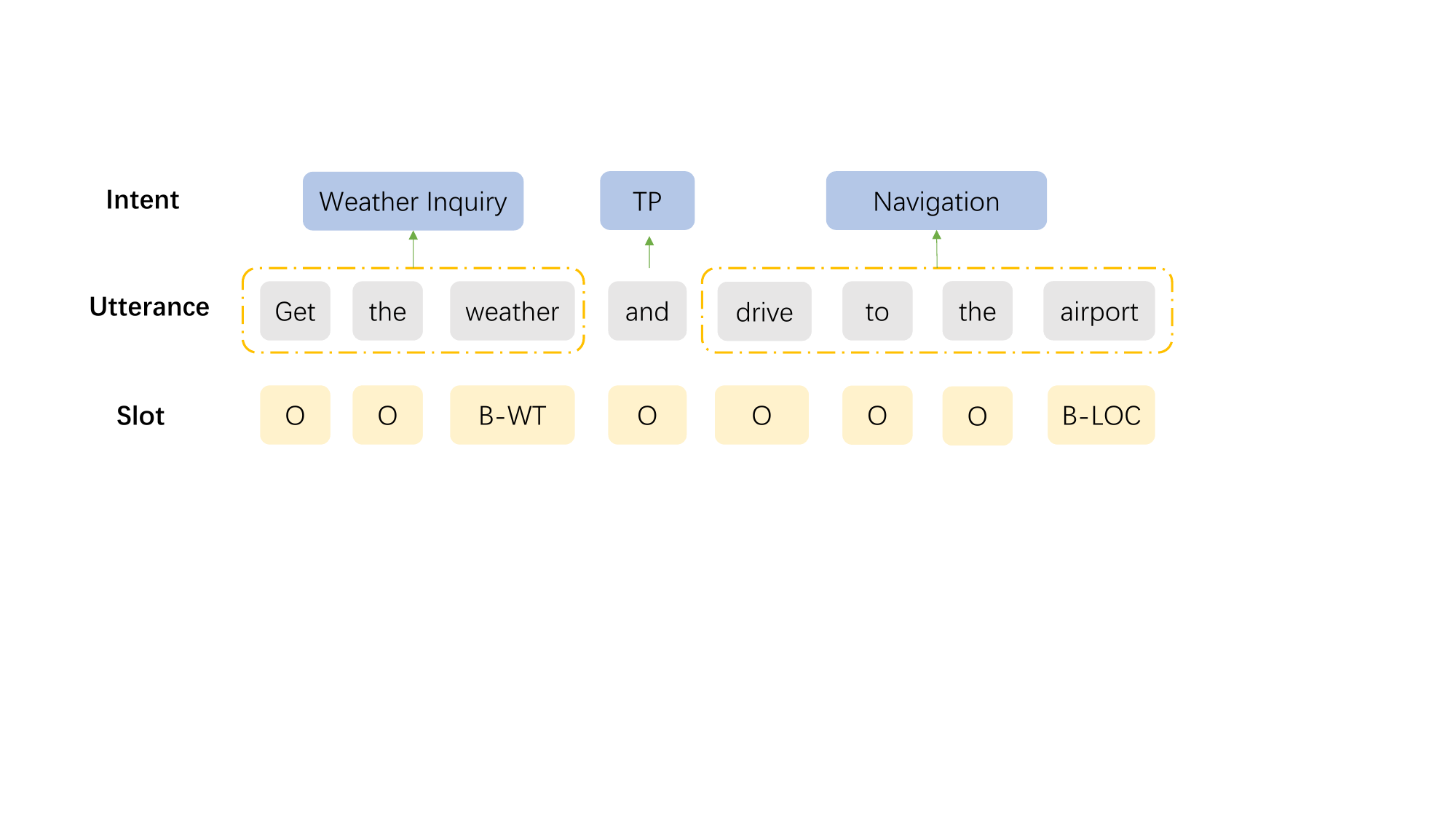}}
    \caption{An example with multi-intent SLU, where B-WT donates B-Weather, B-LOC donates B-Location and “TP” denote “Transition Point”.}
    \label{fig:demo}
\end{figure}

In real-life scenarios, users often express multiple intents within a single utterance, and the Amazon internal dataset showed that 52$\%$ of examples are multi-intent \citep{gangadharaiah2019joint}. Figure \ref{fig:demo} shows a two-intent example, which contains a classification task to classify the intent labels (i.e., predict the intents as : \texttt{Weather\_Inquiry} and \texttt{Navigation}) and a sequence labeling task to predict the slot label sequence (i.e., label the utterance as \{\texttt{O}, \texttt{O}, \texttt{B-WT}, \texttt{O}, \texttt{O}, \texttt{O}, \texttt{O}, \texttt{B-LOC} \}). To deal with multi-intent scenarios, an increasing number of studies have begun to focus on modeling SLU in multi-intent settings. \citet{DBLP:conf/asru/XuS13} and \citet{DBLP:journals/mta/KimRL17} first explored the multi-intent SLU. Then \citet{qin-etal-2020-agif,DBLP:conf/acl/QinWXXCL20} incorporated graph attention networks to model fine-grained intent-slot guiding.
Recently, \citet{DBLP:journals/spl/HuangHZLL22} proposed a chunk-level intent detection (CLID) framework to split multi-intent into single-intent with an intent transition point. Furthermore, \citet{Yin_Huang_Xu_2024} develop an united multi-view intent-slot interaction framework(Uni-MIS), achieving promising performance.

Whether LLMs can effectively handle multi-intent SLU remains an open question. While a straightforward approach might involve fine-tuning LLMs for this specific task, several challenges persist. For example, although LLMs exhibit strong capabilities in entity-level intent detection, their autoregressive architecture can lead to issues such as error propagation and misalignment, particularly in token-level slot filling tasks. This is because LLMs may generate undesirable outputs that do not align one-to-one with the original tokens from the utterance.

To address these challenges, we introduce a novel method that leverages the strengths of LLMs for multi-intent SLU by transforming the traditional token-level slot-filling task into an entity detection problem. By shifting the focus to entity-level slot detection, LLMs can concentrate on identifying relevant slot labels without the need to label every token within a sentence. This approach effectively mitigates the issues of misalignment and uncontrolled generation length.
Moreover, we propose the concept of a {chain of intent}, inspired by the chain-of-thought reasoning framework \cite{chain}. This strategy enhances the ability of LLMs to differentiate and separate multi-intent utterances into distinct sub-intent segments, enabling the models to handle multi-intent recognition in a systematic, step-by-step manner. 

Our experimental results demonstrate that ECLM achieves substantial improvements over state-of-the-art pre-trained models, such as Uni-MIS. Specifically, ECLM achieves overall accuracy gains of 3.7\% on the MixATIS dataset and 3.1\% on the MixSNIPS dataset. Furthermore, the ECLM framework surpasses conventional supervised fine-tuning of LLMs, delivering improvements of 8.5\% and 21.2\% in overall accuracy on MixATIS and MixSNIPS, respectively. In terms of slot filling F1 score, ECLM outperforms vanilla LLM fine-tuning by 22\% and 8.1\%.
We also conduct further experiments to evaluate the performance of ECLM across different numbers of intents within the datasets. Our model consistently outperforms Uni-MIS in overall accuracy across all settings, particularly in scenarios with a high number of intents, showing improvements of 1.1\%, 4.3\%, and 7.8\% for intent counts ranging from 1 to 3. Additionally, we find that ECLM requires only 60\% of the data to surpass Uni-MIS, with more training further enhancing its performance. 
In summary, the contributions of this work can be outlined as follows: (1) We design an entity-slot framework that transforms the traditional token-level slot-filling task into an entity detection problem, thereby mitigating issues of misalignment and uncontrolled generation length. (2) We introduce the {chain of intent} concept, which enables LLMs to effectively handle multi-intent recognition in a step-by-step manner. (3) We demonstrate that our proposed model, {ECLM}, outperforms strong baselines on two widely used datasets, MixATIS and MixSNIPS, across the majority of metrics.

\section{Problem Definition}
\subsection{Multi-Intent Detection}
Given an input sequence $x = (x_1, \ldots, x_n)$, multi-intent detection can be defined as a multi-label classification task that outputs a sequence of intent labels $o_I = (o_{1}^I, \ldots, o_{m}^I)$, where $m$ is the number of intents in a given discourse and $n$ is the length of the discourse.

\begin{figure*}[t] 
    \centering  
    \centerline{\includegraphics[width=0.9\linewidth]{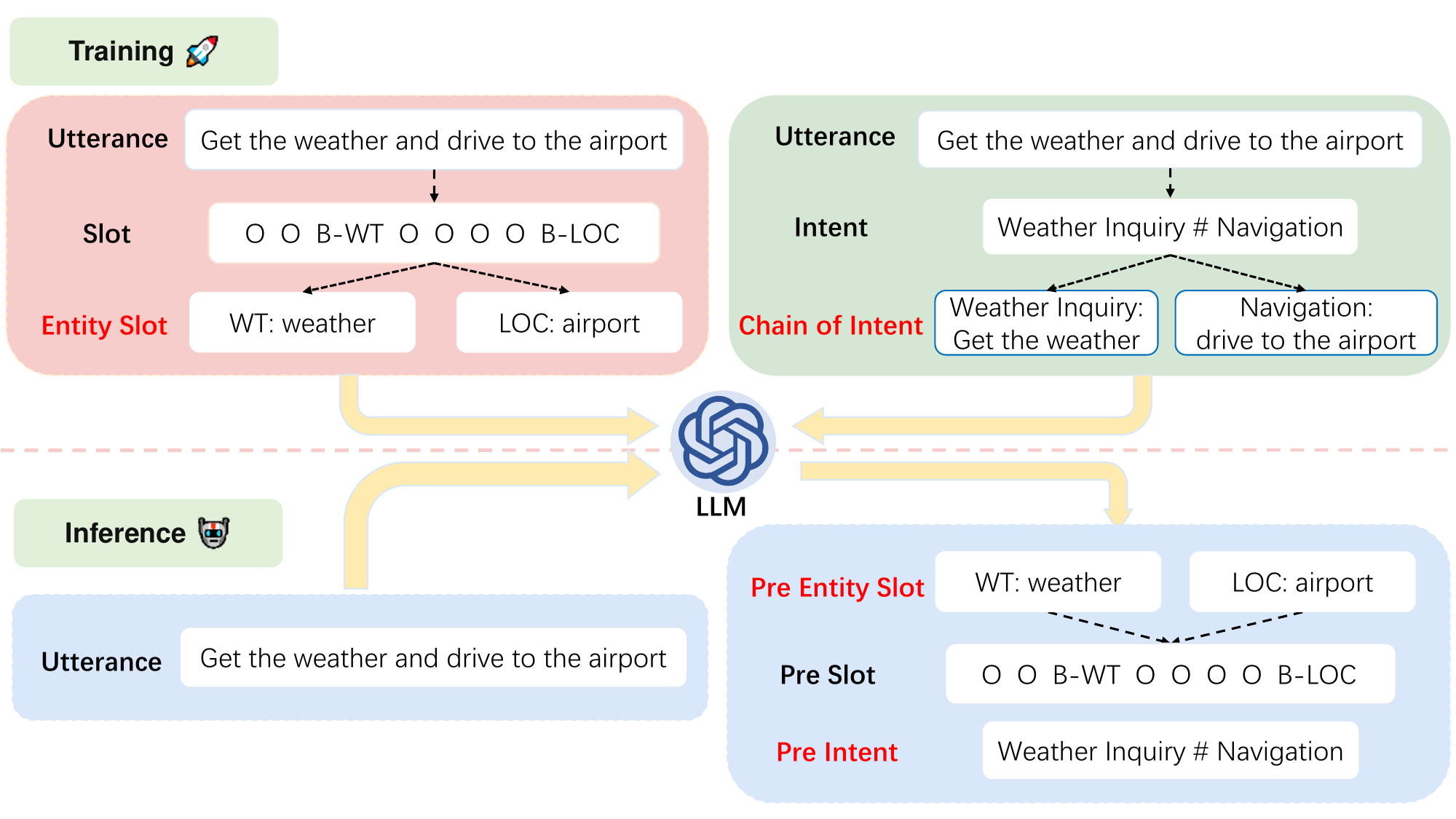}}
   \caption{The 
   key components of the ECLM and the different operations performed by the same example in the training phase as well as in the inference phase.}
    \label{Core}
\end{figure*}

\subsection{Slot Filling}
Slot filling can be considered as a sequence annotation task that maps the input discourse $x$ to a slot output sequence $o_S = (o_{1}^S, \ldots, o_{n}^S)$.

\section{Approach}

\begin{figure*}[ht] 
    \centering  
    \centerline{\includegraphics[width=0.9\linewidth]{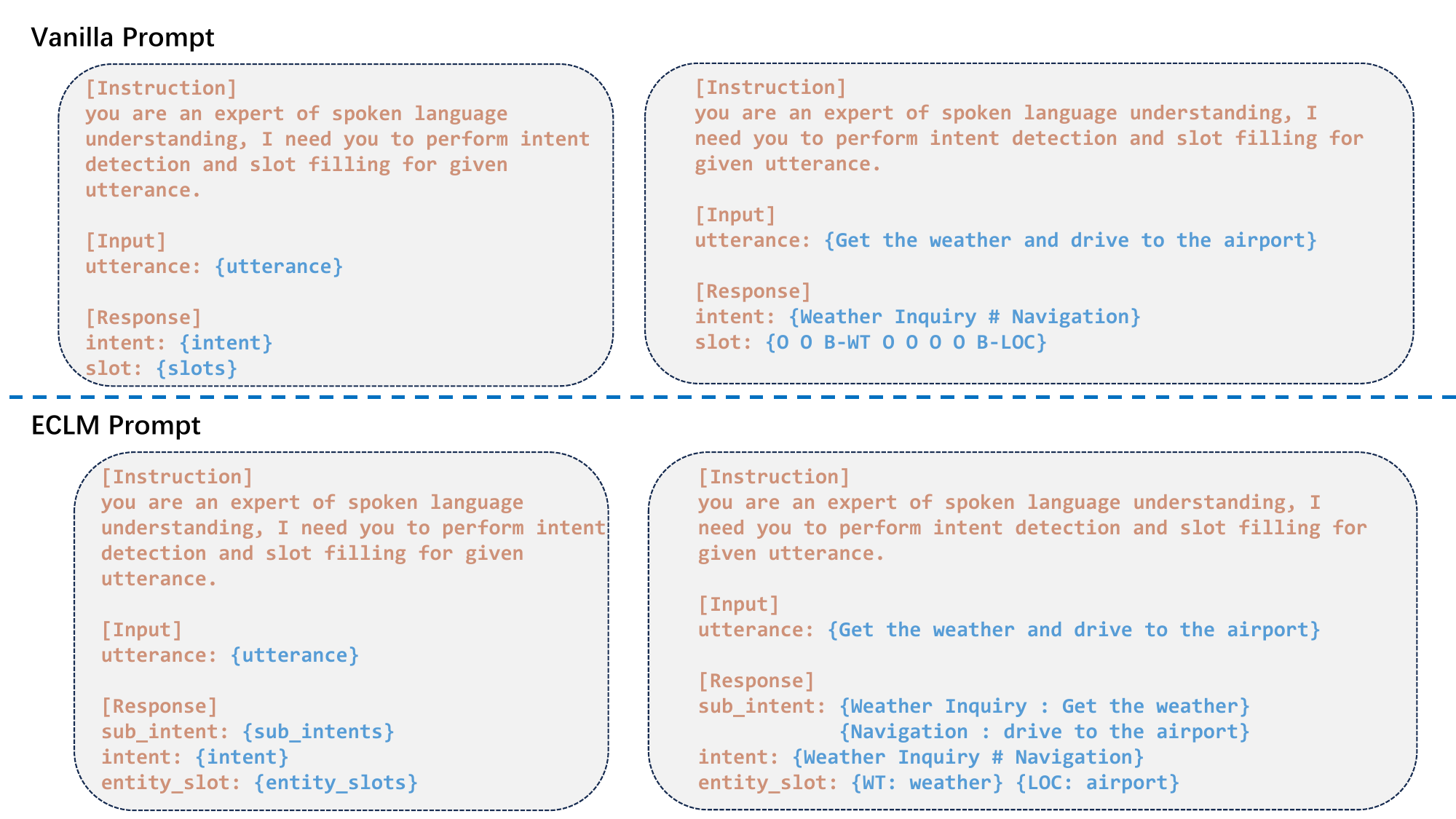}}
    \caption{Comparison of prompt structures used in ECLM versus Vanilla SFT.}
    \label{Prompt}
\end{figure*}

As shown in Figure \ref{Core}, our approach establishes a comprehensive framework for integrating LLMs into the domain of multi-intent SLU. By showing an example of the ECLM training process, the key components of the framework are highlighted: the Entity Slots and the Chain of Intent. Finally, we perform supervised fine-tuning to adapt the LLM to the multi-intent SLU task. Detail information of the prompt template can be seen in the Figure \ref{Prompt}.





\subsection{Entity Slots Construction and Recovery}

Our approach introduces a novel two-phase process: Entity Slots Construction for training, and Entity Slots Recovery for inference, designed to bridge the gap between traditional sequence labeling and the generative capabilities of LLMs.

\subsubsection{Entity Slots Construction}

In the Entity Slots Construction phase, we transform conventional BIO sequence labeling into a structured entity-slot representation, optimizing for generative modeling with LLMs. Given a token sequence \( T = \{t_1, t_2, \dots, t_n\} \) and its corresponding BIO-annotated tags \( S = \{s_1, s_2, \dots, s_n\} \), we map these to a set of entity slots \( E = \{e_1, e_2, \dots, e_m\} \), where \( m \) is the number of identified entities. This mapping is defined by a function \( c \) as follows:
\begin{equation}
    E = c(T, S) = \left\{ \left( k_i, \textstyle\bigcup_{j \in I_i} t_j \right) \right\}_{i=1}^{m},
\end{equation}
where \( k_i \) is the entity type extracted from the prefix of the 'B-' tag, and \( I_i \) is the index set of tokens that belong to the \(i\)-th entity. Each entity starts with a token labeled 'B-XXX' and includes all consecutive tokens labeled 'I-XXX' of the same type. New entities are initiated by a new 'B-' tag or interrupted by an 'O'. This function systematically extracts and groups contiguous tokens belonging to each entity, ensuring they are correctly concatenated to form complete slot values.


\subsubsection{Entity Slots Recovery}

During the inference stage, we implement an Entity Slots Recovery process to convert the generated structured entity slots back into a BIO-tagged sequence. This recovery process, defined by a function \( r \), can be expressed as:

\begin{equation}
    r(T, E) = \{s_j\}_{j=1}^{n} ,
\end{equation}
where \( s_j \) is determined for each token \( t_j \) based on its presence in the entity slots \( E \). The recovery follows these rules:
(1) If \( t_j \) is the first token of an entity in \( E \), \( s_j \) is assigned a 'B-' tag with the corresponding entity type.
(2) If \( t_j \) is a non-initial token of an entity in \( E \), \( s_j \) is assigned an 'I-' tag with the corresponding entity type.
(3) If \( t_j \) does not belong to any entity in \( E \), \( s_j \) is assigned an 'O' tag.


\subsection{Chain of Intent}

To effectively manage the complexity of multi-intent SLU, we propose a novel framework termed the "Chain of Intent," inspired by the "Chain of Thought" reasoning process \cite{chain}. This framework enhances the model's ability to discern and process multiple intents within a single utterance by segmenting it into distinct sub-intent utterances, enabling more granular understanding and response generation.

Consider an utterance \( U \) consisting of \( n \) intents. Each intent \( I_i \) (where \( i = 1, 2, \dots, n \)) corresponds to a specific segment of the utterance \( U_i \). The process of decomposing the utterance \( U \) can be formally expressed as a mapping:
\begin{equation}
    U \mapsto \{(I_1: U_1), (I_2: U_2), \dots, (I_n: U_n)\},
\end{equation}
where the structured pairs \( (I_i: U_i) \) represent each intent \( I_i \) paired with its associated sub-utterance \( U_i \). During training, the model is presented with this mapping to learn the relationship between each intent and its corresponding segment of the utterance, thereby improving its ability to generate contextually accurate and intent-specific responses.

\subsection{Supervised Fine-tuning}

We employ supervised fine-tuning to enhance the generative capabilities of LLMs, ensuring they meet the structured requirements of multi-intent spoken language understanding (SLU). This process involves adjusting the model parameters \( \theta \) to minimize a loss function \( \mathcal{L} \) across a set of training examples.
Given a training set \( \{(U_j, T_j)\}_{j=1}^M \), where \( U_j \) represents the \( j \)-th input utterance and \( T_j \) denotes the corresponding target output, including segmented sub-intents and entity slots, the fine-tuning objective is defined as:
\begin{equation}
    \theta^* = \arg\min_{\theta} \sum_{j=1}^{M} \mathcal{L}\left(\text{LLM}(U_j; \theta), T_j\right),
\end{equation}
where, \( \text{LLM}(U_j; \theta) \) represents the output generated by the LLM given the input \( U_j \) with parameters \( \theta \). The supervised fine-tuning process iteratively updates \( \theta \) to more accurately map input utterances \( U_j \) to their corresponding intent and entity slot outputs \( T_j \), thereby improving the model's effectiveness in multi-intent SLU tasks.

\section{Experiments}


\subsection{Datasets}

We conducted experiments on two widely used multi-intent SLU datasets: MixATIS \cite{DBLP:conf/naacl/HemphillGD90,qin-etal-2020-agif} and MixSNIPS \cite{origin_snips,qin-etal-2020-agif}. The MixATIS dataset contains 13,162 training instances and 828 test instances, primarily focusing on airline-related queries. In contrast, the MixSNIPS dataset spans a broader range of domains, including restaurants, hotels, and movies, with 39,776 training instances and 2,199 test instances. These datasets are designed to mimic real-world scenarios, featuring utterances with 1 to 3 intents, distributed in ratios of 30\%, 50\%, and 20\%, respectively and detail information can be found in Table \ref{dataset}.

\subsection{Experimental Settings}
We use LLaMA 3.1–8B-Instruct as the base model and conduct our experiments with a carefully tuned set of hyperparameters. Additionally, we evaluate the performance of different backbone models, as shown in Table~\ref{backbone_table}.
To determine the optimal settings, we performed a grid search over the learning rate \([1 \times 10^{-5}, 2 \times 10^{-5}, 5 \times 10^{-5}, 1 \times 10^{-4}]\) and the number of epochs \([1, 2, 3]\). Based on the results, we settled on a learning rate of \(2 \times 10^{-5}\) and a batch size of 32, tuning the model for 1 epoch on both datasets.
During inference, a generation temperature of 0.0 was used to ensure deterministic and consistent outputs.

\begin{table}[h]
\centering
\resizebox{0.9\columnwidth}{!}{
\begin{tabular}{lrrr}
\toprule
Dataset                 & MixATIS      & MixSNIPS     \\
\midrule
Vocabulary Size         & 722       & 11241     \\
Intent categories             & 17        & 6         \\
Slot categories              & 116       & 71        \\
Training set size            & 13162      & 39776     \\
Test set size                 & 828       & 2199       \\
\bottomrule
\end{tabular}
}
\caption{Dataset statistics}
\label{dataset}
\end{table}

\begin{table*}[t]
 \begin{center}
 \scalebox{0.85}{
 \begin{tabular}{l|ccc|ccc}
  \hline
 \multicolumn{1}{l|}
{\multirow{2}{*}{\bf Model}} & 
\multicolumn{3}{c|}{\bf MixATIS } & \multicolumn{3}{c}{ \bf MixSNIPS }  \\
\cline{2-7} 
\multicolumn{1}{c|}{}  & \multicolumn{1}{c}{Slot(F1)} & \multicolumn{1}{c}{Intent(Acc)} & \multicolumn{1}{c|}{Overall(Acc)}  & \multicolumn{1}{c}{Slot(F1)} & \multicolumn{1}{c}{Intent(Acc)} & \multicolumn{1}{c}{Overall(Acc)}\\
\cline{1-7} 
Stack-Propagation \citep{DBLP:conf/emnlp/QinCLWL19}       
& 87.8 & 72.1   & 40.1       & 94.2 & 96.0 & 72.9   \\
AGIF   \citep{DBLP:conf/emnlp/QinXCL20}  
 & 86.9 & 72.2  & 	39.2 & 93.8  & 95.1  & 72.7 \\
GL-GIN  \citep{DBLP:conf/acl/QinWXXCL20}       
& 87.2 & 75.6    & 41.6  & 93.7 & 95.2   & 72.4  \\
SDJN  \citep{DBLP:conf/icassp/ChenZZ22}       
& 88.2 & 77.1    & 44.6 & 94.4 & 96.5   & 75.7  \\

CLID  \citep{DBLP:journals/spl/HuangHZLL22}       
& 88.2 & 77.5    & 49.0 & 94.3 & 96.6   & 75.0  \\



SSRAN \cite{DBLP:conf/aaai/ChengY023} & {89.4}   &  77.9   &  48.9 & 95.8   &  {98.4} &   77.5   \\
 \hline
SDJN + Bert
& 87.5 & {78.0}    & 46.3 & 95.4 & 96.7   &79.3  \\

RoBERTa+Linear
&  86.0  & 80.3  & 48.4   &  96.0  & 97.4  & 82.1  \\

CLID + Roberta
& 85.9 & 80.5    & 49.4 & 96.0 & 97.0   & 82.2\\

Uni-MIS \cite{Yin_Huang_Xu_2024}
& {88.3} & {78.5}  &  {52.5}    &    {96.4} & {97.2} & {83.4} \\








\hline



\hline

ECLM (Ours)
& \textbf{90.2} & \textbf{80.7} &  $\textbf{56.2}^*$  &    \textbf{97.0} & {97.0} & $\textbf{86.5}^*$ \\ 
\hline

\end{tabular}
}
\end{center}
 \caption{
Multi-Intent SLU performance on MixATIS and MixSNIPS datasets.
Values with * indicate that the improvement from our model is statistically significant over all baselines ($p < 0.05$ under t-test).}
\label{main_table} 
\end{table*}

\begin{table*}[t]
 \begin{center}
 \scalebox{0.95}{
 \begin{tabular}{l|ccc|ccc}
  \hline
 \multicolumn{1}{l|}
{\multirow{2}{*}{\bf Model}} & 
\multicolumn{3}{c|}{\bf MixATIS Dataset}  & \multicolumn{3}{c}{ \bf MixSNIPS Dataset} \\
\cline{2-7} 
\multicolumn{1}{c|}{}  &
\multicolumn{1}{c}{Slot(F1)} &
\multicolumn{1}{c}{Intent(Acc)} & \multicolumn{1}{c|}{Overall(Acc)} &
\multicolumn{1}{c}{Slot(F1)} & 
\multicolumn{1}{c}{Intent(Acc)} & \multicolumn{1}{c}{Overall(Acc)}\\
\cline{1-7} 
ECLM (Ours) &  90.2 & 80.7 & \textbf{56.2}  &    {97.0} & {97.0} & \textbf{86.5} \\ 
\hline

-w/o  Entity Slot    & 73.5 & 78.7 & 54.9   & 92.7 & 97.6 &69.7 \\

-w/o Chain of Intent
& 89.4 & 82.6 & 52.9    &   96.8 & 98.0 &85.1 \\ 

-w/o Both (Vanilla SFT)
& 68.2 & 74.0 &  47.7   &   88.9 & 97.4 & 65.3 \\ 

\hline
 \end{tabular}
 }
\caption{Ablation experiments on the MixATIS and MixSNIPS datasets. Interestingly, we observe that entity slots play a more significant role in the MixSNIPS dataset compared to MixATIS, while the chain of intent does not explicitly improve intent accuracy but instead enhances overall performance.}

  \label{ablation}
 \end{center} 
\end{table*}

\subsection{Baselines}


In our study, we benchmark LLMs performance against a range of established baselines in the multi-intent SLU domain. These include \textbf{vanilla models} like Stack-Propagation \citep{DBLP:conf/emnlp/QinCLWL19}: a stack-propagation framework to explicitly incorporate intent detection for guiding slot filling. AGIF \citep{DBLP:conf/emnlp/QinXCL20}: an adaptive interaction network to achieve fine-grained multi-intent information integration, GL-GIN \citep{DBLP:conf/acl/QinWXXCL20}: a local slot-aware and global intent-slot interaction graph framework to model the interaction between multiple intents and all slots within an utterance, SDJN \citep{DBLP:conf/icassp/ChenZZ22}: a multiple instance learning and self-distillation framework for weakly supervised multiple intent information capturing, CLID \citep{DBLP:journals/spl/HuangHZLL22}:  a chunk-level intent detection framework for recognizing intent within a fragment of an utterance and SSRAN \citep{DBLP:conf/aaai/ChengY023}: a transformative network built on the Transformer model, designed to reduce the complexity of multi-intent detection in SLU through scope recognition and bidirectional interaction between results of slot filling and intent detection. We also included \textbf{PLM-based models} such as Uni-MIS \citep{Yin_Huang_Xu_2024}: a unified multi-intent slu framework via multi-view intent-slot interaction. Additionally, SDJN(Bert) and CLID(Roberta) extend their respective base models by incorporating pre-trained language model backbones.





\begin{table*}[ht]
 \begin{center}
 \scalebox{0.7}{
\begin{tabular}{l|ccc|ccc|ccc}
\hline
\multicolumn{1}{l|}
{\bf{Model}} & 
\multicolumn{3}{c|}{\bf intent num = 1} & \multicolumn{3}{c|}{ \bf intent num = 2} & \multicolumn{3}{c}{ \bf intent num = 3}\\
\cline{2-10} 
\multicolumn{1}{c|}{}  & \multicolumn{1}{c}{Slot(F1)} & \multicolumn{1}{c}{Intent(Acc)} & \multicolumn{1}{c|}{Overall(Acc)}& \multicolumn{1}{c}{Slot(F1)}   & \multicolumn{1}{c}{Intent(Acc)}  & \multicolumn{1}{c|}{Overall(Acc)}  & \multicolumn{1}{c}{Slot(F1)} & \multicolumn{1}{c}{Intent(Acc)} & \multicolumn{1}{c}{Overall(Acc)} \\
\cline{1-10} 

GL-GIN  & 88.0 & 91.3  & 72.6   & 87.3 & 76.2 & 39.1  & 86.8 & 63.1 & 23.0\\
CLID    & 88.6 & 94.7  & 76.4   & 88.1 & 77.5 & 48.4  & 87.6 & 64.3 & 28.5 \\

\hline
CLID + Roberta    & 88.6 & 95.8 &  77.6  & 85.4 & 80.3 & 48.8  & 84.7  & 66.8 & 29.0 \\

Uni-MIS &  89.2   &  95.1   &  78.6 &  87.6   &  78.3 &   50.5   &  86.7   &  66.7  &  {31.7} \\

\hline




ECLM(Ours)  &  92.1   & 93.7   & \textbf{79.7} &  90.3  &  79.4  &   \textbf{54.8}   &  90.3   &  70.0  & \textbf{39.5} \\

\hline
\end{tabular}
}
\end{center}
\caption{The result comes from the dataset MixATIS. The {intent num} denotes the number of intents in an utterance. 
}
\label{split}
\end{table*}

\begin{figure*}[ht]
    \centering
    \subfigure[ MixATIS dataset]{\includegraphics[width=0.38\textwidth]{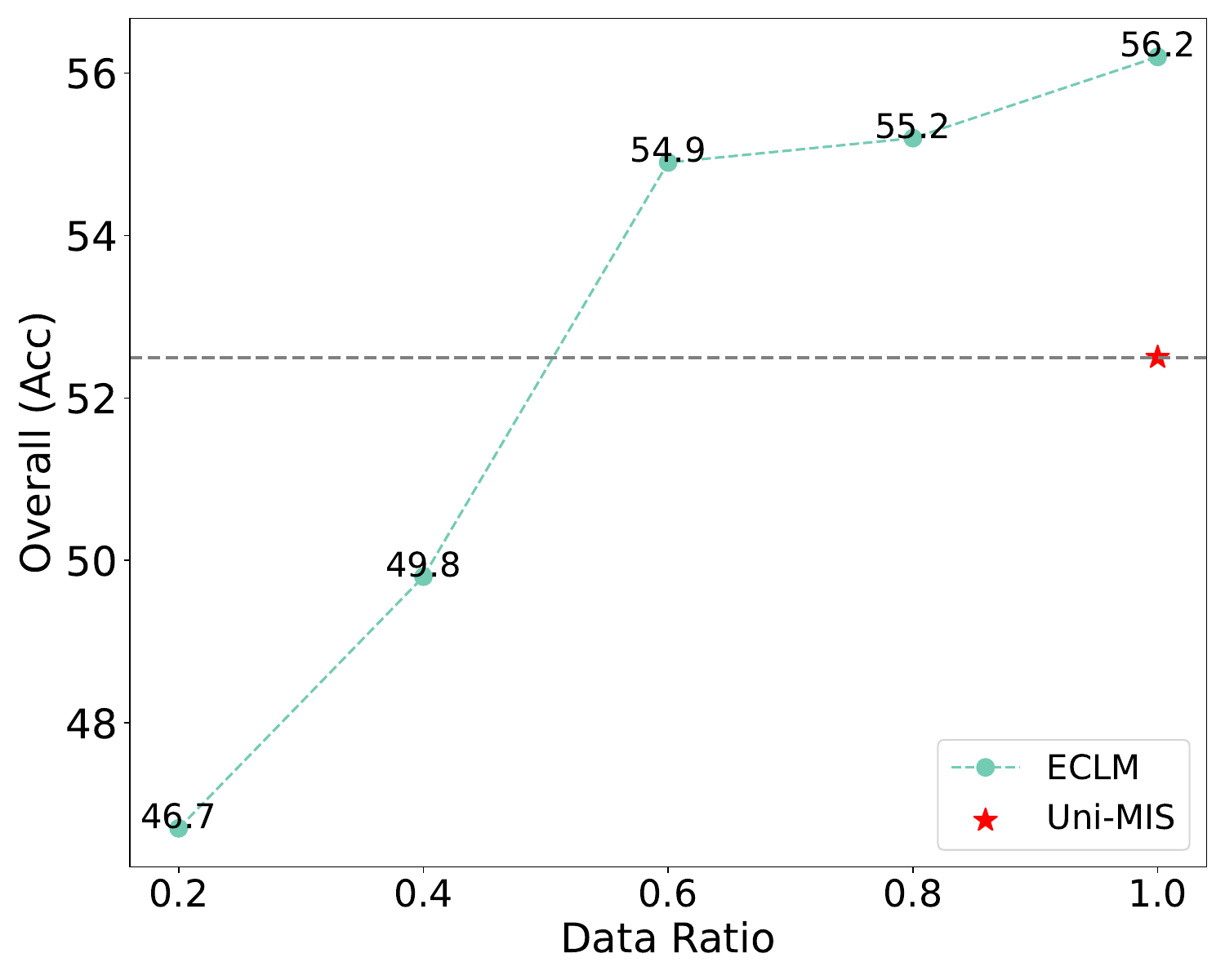}
    }
   \centering
    \subfigure[MixSNIPS dataset]{
\includegraphics[width=0.38\textwidth]{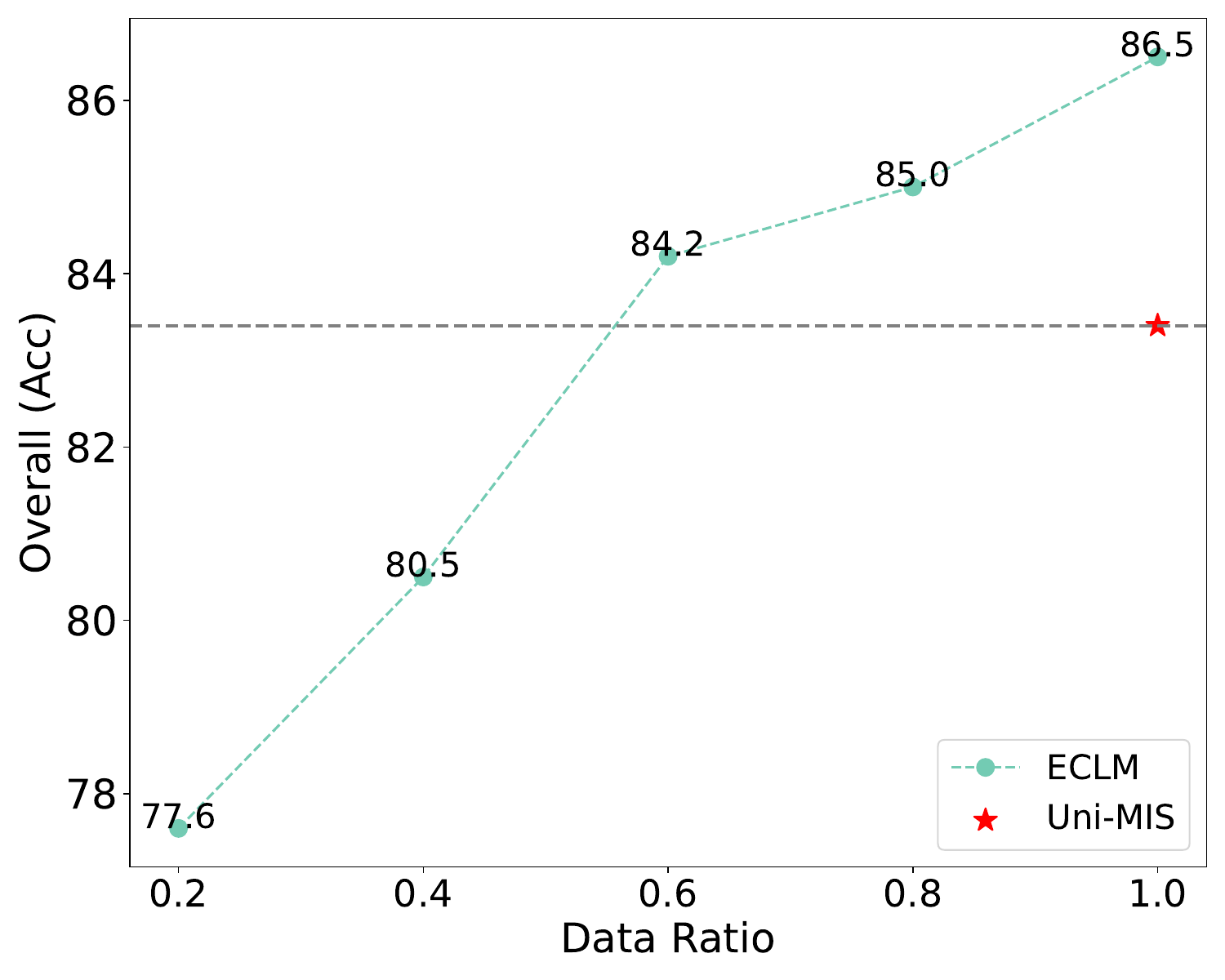}
}
    \caption{Performance of ECLM  on the MixATIS and MixSNIPS datasets at different training data proportions}
    \label{fig:data_rate}
\end{figure*}

\begin{figure*}[t] 
    \centering  
    \centerline{\includegraphics[width=.90\linewidth]{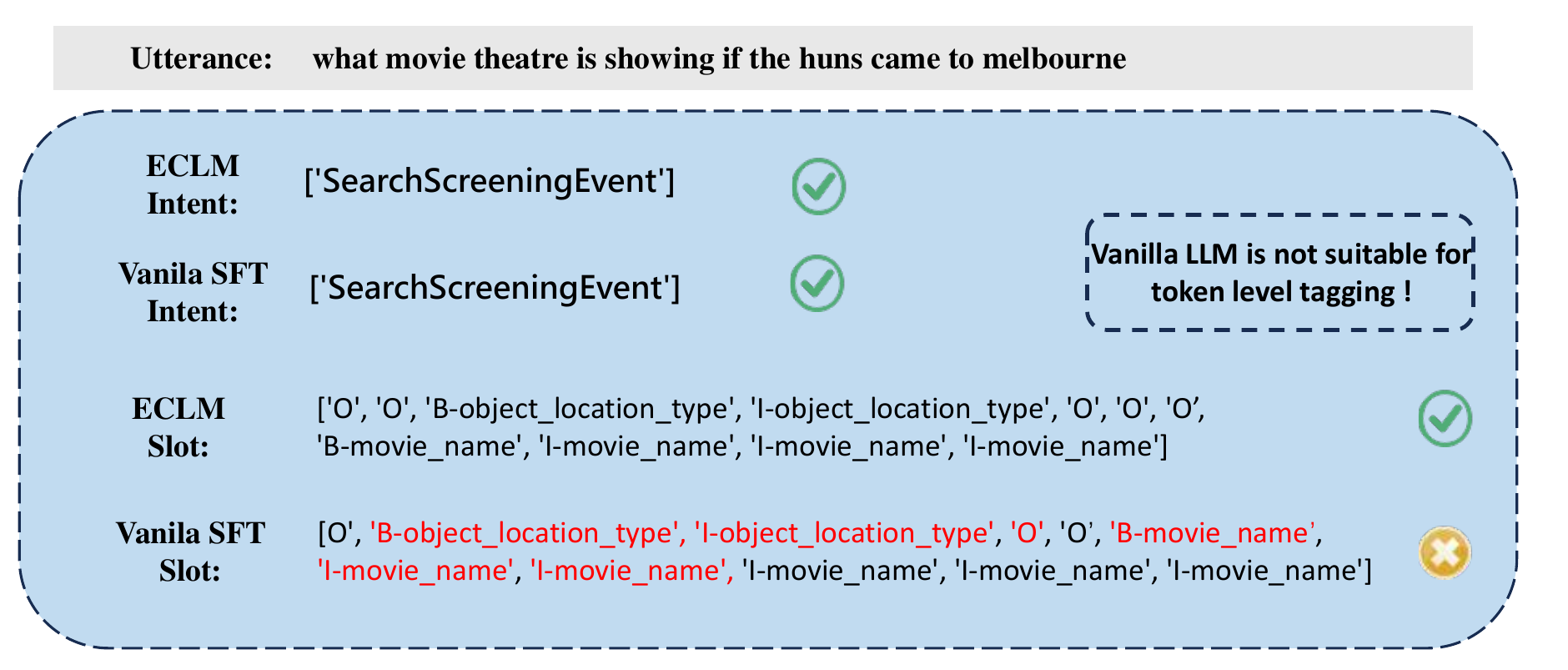}}
    \caption{Comparative analysis of ECLM and vanilla SFT performance on a complex multi-intent utterance, highlighting ECLM's superior slot filling capabilities and the limitations of LLMs in token-level tagging tasks. More case can refer to the Section \ref{Case} in Appendix.}
    \label{main_case}
\end{figure*}

\subsection{Main Result Analysis}
The evaluation metrics included slot F1 score, intent accuracy and semantic accuracy to comprehensively assess the sentence-level semantic frame parsing capabilities. These metrics, adhering to the methodologies delineated by \citet{DBLP:conf/acl/QinWXXCL20,DBLP:journals/spl/HuangHZLL22,Yin_Huang_Xu_2024} facilitate a nuanced evaluation of SLU systems. The paramount metric, semantic overall accuracy, quantifies the system's proficiency in simultaneously and correctly predicting both intents and slots within a single sentence. 

Our main experiments yield several important observations:
(1) As shown in Table \ref{main_table}, ECLM outperforms the strong baseline in slot filling F1 scores in both datasets. This improvement indicates that the ECLM interaction effectively utilises entity slots to improve it's slot filling ability.
(2) For the single-domain MixATIS dataset, ECLM outperforms Uni-MIS with a 1.9 \% point improvement in slot filling F1 scores (90.2\%), a 2.2 \% point improvement in intent prediction accuracy (80.7\%), and a 3.7 \% point improvement in overall sentence-level semantic frame parsing accuracy (56.2\%).
For the multi-domain MixATIS dataset, ECLM outperforms Uni-MIS by 0.6 \% points in slot-filling F1 score (97.0\%) and 3.1 \% points in overall sentence-level semantic frame parsing accuracy (86.5\%). These results highlight the competitive advantage of robust language models in multi-intent SLU tasks.
(3) Importantly, our framework achieves state-of-the-art performance for most evaluation metrics, highlighting a promising research direction for multi-intent SLU using LLM-based methodologies.

\subsection{Ablation Study}

To understand the impact of key components in ECLM, we conducted ablation experiments on the MixATIS and MixSNIPS datasets. As shown in Table \ref{ablation}, the results illustrate the contribution of entity slots and the chain of intent to overall performance.

\subsubsection{Without Entity Slot}
Removing the entity slot significantly reduces performance, with a drop of 16.7 \% in slot F1 score and 1.3 \% points in overall accuracy on MixATIS. Similarly, on MixSNIPS, we observe a drop of 4.3 \% in slot F1 score, and the overall accuracy decreases by 16.8 \%. This highlights the crucial role of entity slots in maintaining high performance. Especially in the multi-domain dataset MixSNIPS, the absence of entity slots may cause significant misalignment, as the majority of slot labels are "O". This could lead to the model incorrectly labeling words as "O" rather than their corresponding slot tags.

\subsubsection{Without Chain of Intent}
Eliminating the chain of intent structure leads to a 0.8 \% point drop in slot F1 score  and a 3.3 \% decline in overall accuracy  on MixATIS. On MixSNIPS, the overall accuracy decreases by 1.4 \%, emphasizing the importance of intent chaining in enhancing the model's semantic understanding. However, we observe that the improvement in intent detection accuracy is less pronounced, suggesting that the chain of intent mainly contributes to the joint effect and compromises some intent accuracy.

\subsubsection{Without Both (Vanilla SFT)}
When both components are removed, the performance suffers dramatically. The slot F1 score drops by 22.0 \% and the overall accuracy by 8.5 \% on MixATIS. The MixSNIPS dataset also shows a significant decrease, with the overall accuracy dropping by 21.2 \%. This indicates that the Vanilla SFT method cannot effectively adapt LLMs to this domain.






\section{Further Exploration}

\subsection{Influence of Different Intent Numbers}

The analysis of MixATIS dataset results, categorized by the number of intents as shown in Table \ref{split}, reveals significant insights into the performance of our ECLM model compared to baseline approaches. For single-intent utterances, ECLM achieves superior performance with a slot F1 score of 92.1\% and overall accuracy of 79.7\%, outperforming the strong Uni-MIS over Uni-MIS (89.2\% and 78.6\% respectively). As the complexity increases with multi-intent scenarios, ECLM's advantages become more pronounced. In two-intent cases, ECLM maintains its lead with a slot F1 of 90.3\% and overall accuracy of 54.8\%, showing a substantial improvement over Uni-MIS (87.6\% and 50.5\% respectively). The performance gap widens further for three-intent utterances, where ECLM achieves a slot F1 of 90.3\%, intent accuracy of 70.0\%, and overall accuracy of 39.5\%, significantly surpassing Uni-MIS (86.7\%, 66.7\%, and 31.7\% respectively). This consistent outperformance, particularly in challenging multi-intent scenarios, underscores ECLM's robustness and efficacy in handling complex spoken language understanding tasks. The results demonstrate ECLM's capacity to maintain high performance across varying levels of intent complexity, indicating its potential as a versatile solution for advanced SLU systems.

\subsection{Influence of Training Data Ratio}

Figure \ref{fig:data_rate} illustrates the impact of varying training data volumes on ECLM's performance, focusing on overall semantic accuracy across the MixATIS and MixSNIPS datasets. We systematically adjusted the training data ratios at 0.2, 0.4, 0.6, 0.8, and 1.0 to assess model proficiency under different data availability scenarios.
The results demonstrate a consistent positive correlation between the data ratio and performance improvements across both datasets. For MixATIS, ECLM's semantic accuracy rises from 46.7\% at 0.2 data ratio to 56.2\% at full data utilization, surpassing the Uni-MIS baseline (52.5\%) with just 60\% of the training data. Similarly, on MixSNIPS, ECLM's performance increases from 77.6\% to 86.5\%, exceeding the Uni-MIS benchmark (83.4\%) also at approximately 60\% data ratio.
Notably, ECLM exhibits robust performance even with limited data, achieving competitive results at lower data ratios. The performance gains are more pronounced in the MixSNIPS dataset, suggesting ECLM's particular effectiveness in multi-domain scenarios. As the data ratio approaches 1.0, the performance improvement rate gradually stabilizes, indicating a potential plateau effect at higher data volumes.

\subsection{Influence of Different Backbone LLMs in the ECLM Framework}

Table \ref{backbone_table} presents a comparative analysis of overall accuracy across various LLMs when integrated into our ECLM framework, evaluated on both the MixATIS and MixSNIPS datasets. The results demonstrate a clear progression in performance as we move towards more advanced LLM architectures.
Llama2-7B-Chat, while competent, shows the lowest performance with overall accuracies of 48.2\% and 81.5\% on MixATIS and MixSNIPS respectively. Mistral-7B-Instruct-v0.1 exhibits a notable improvement, achieving 50.1\% and 83.9\% on the same datasets, highlighting the rapid advancements in LLM capabilities.
The Llama3.1 series showcases significant performance gains. The base Llama3.1-8B model achieves impressive results of 55.6\% and 85.9\% on MixATIS and MixSNIPS, respectively. However, the instruction-tuned variant, Llama3.1-8B-Instruct, emerges as the top performer, reaching 56.2\% accuracy on MixATIS and 86.5\% on MixSNIPS.
The superior performance of Llama3.1-8B-Instruct underscores the importance of instruction tuning in enhancing model capabilities for specific tasks like multi-intent SLU. This model's consistent outperformance across both datasets justifies its selection as the default backbone for our ECLM framework.

\begin{table}[t]
 \begin{center}
 \scalebox{0.85}{
 \begin{tabular}{l|cc}
  \hline
 \multicolumn{1}{l|}
{\bf Model} & 
\multicolumn{1}{c}{\bf MixATIS} & 
\multicolumn{1}{c}{\bf MixSNIPS} \\
\cline{1-3} 

Llama2-7B-Chat     
 & 48.2 & 81.5 \\

Mistral-7B-Instruct-v0.1      
 & 50.1 & 83.9 \\

Llama3.1-8B
 & 55.6 & 85.9 \\
\hline

Llama3.1-8B-Instruct  
 & \textbf{56.2} & \textbf{86.5} \\ 
\hline

\end{tabular}
}
\end{center}
\caption{The impact of different backbone LLMs Integrated into the ECLM Framework.}
\label{backbone_table} 
\end{table}

\subsection{Case Analysis}
As illustrated in Figure \ref{main_case}, we present a comparative analysis of ECLM and vanilla LLM-based SFT approaches on a complex multi-intent utterance. The example, "what movie theatre is showing if the huns came to melbourne", demonstrates the superior performance of ECLM in handling intricate spoken language understanding tasks.
Both ECLM and vanilla SFT correctly identify the primary intent as "SearchScreeningEvent". However, the critical distinction emerges in the slot filling task. ECLM accurately labels each token, precisely identifying "movie theatre" as the "object\_location\_type" and "if the huns came to melbourne" as the "movie\_name". In contrast, the vanilla SFT model, despite its correct intent classification, exhibits significant errors in slot filling.
The vanilla SFT incorrectly labels "what" as part of the "object\_location\_type" and mistakenly extends the "movie\_name" to include "showing". This misalignment highlights a fundamental limitation of autoregressive LLMs in token-level tagging tasks. The sequential nature of their predictions can lead to error propagation and misalignment with the original utterance tokens.

\section{Related Work}
\label{sec:appendix}

\subsection{Intent Detection and Slot Filling}

The inherent interconnected of intent detection and slot filling has spurred the development of unified models that foster mutual interaction between the two elements. Joint learning techniques, acknowledging the potent correlation between intents and slots, have proven particularly efficacious in recent years. Certain methodologies facilitating simultaneous slot filling and intent detection employ shared parameters \citep{DBLP:conf/interspeech/LiuL16,DBLP:conf/ijcai/ZhangW16a,DBLP:conf/naacl/WangSJ18}, while others model the relationship between the two via either unidirectional interaction or bidirectional-flow interaction \citep{DBLP:conf/ijcai/QinXC021}.
Models adopting unidirectional interaction, such as those by \citep{DBLP:conf/naacl/GooGHHCHC18, DBLP:conf/emnlp/LiLQ18, DBLP:conf/emnlp/QinCLWL19}, primarily emphasize the flow from intent to slot. Gating mechanisms, functioning as specialized guiding forces for slot filling, have seen extensive use \citep{DBLP:conf/naacl/GooGHHCHC18, DBLP:conf/emnlp/LiLQ18}. \citet{DBLP:conf/emnlp/QinCLWL19} put forth a token-level intent detection model to curtail error propagation.
Bidirectional-flow interaction models \citep{DBLP:conf/acl/ENCS19, DBLP:conf/acl/ZhangLDFY19, DBLP:conf/emnlp/LiuMZZCX19, DBLP:conf/icassp/QinLCKZ021}, on the other hand, examine the reciprocal influence of intent detection and slot filling. \citet{DBLP:conf/acl/ENCS19} utilized iterative mechanisms to enhance intent detection and slot filling in both directions. 
Fine-grained intent detection and intent-slot interaction models have also seen remarkable advancements. 
\citet{DBLP:conf/icassp/ChenZZ22} developed a Self-distillation Joint SLU model exploitating multi-task learning, and treated multiple intent detection as a weakly-supervised problem solved through Multiple Instance Learning (MIL). Similarly, \citet{DBLP:journals/spl/HuangHZLL22} introduced a chunk-level intent detection framework that employs an auxiliary task to pinpoint intent transition points within utterances, thereby augmenting the recognition of multiple intents. Furthermore, \citet{DBLP:conf/aaai/ChengY023} proposed a transformative network rooted in the Transformer model, designed to diminish the complexity of multi-intent detection in SLU. Recently, \citet{Yin_Huang_Xu_2024} further develop an united multi-view intent-slot interaction framework(Uni-MIS), archiving promising performance.

\subsection{Open Source LLMs}

The advent of open-source LLMs such as Llama \citep{llama2}, Vicuna \cite{vicuna}, and Mistral \cite{mistral} has dramatically reshaped the landscape of NLP. These models, characterized by their vast parameter spaces and diverse training corpora, have significantly expanded the capabilities and applications of NLP technologies. The rapid evolution of LLMs has accelerated progress across a broad spectrum of NLP tasks, including natural language inference, summarization, and dialogue systems \cite{Geogle,DBLP:conf/eacl/KavumbaBHI23}. 
Complementing these advancements, the "Chain of Thought" method \cite{chain} has emerged as a pivotal technique in enhancing the reasoning capabilities of LLMs. This approach enables models to break down complex problems into interpretable steps, significantly improving performance on tasks requiring multi-step reasoning or complex problem-solving. 

\section{Conclusion}
In this paper, we introduced the Entity-level Large Language Model framework ECLM for multi-intent spoken language understanding. By transforming token-level slot-filling into an entity recognition problem and introducing the "Chain of Intent" concept, we effectively addressed the challenges of applying LLMs to SLU tasks. Our approach significantly outperformed state-of-the-art models, including Uni-MIS and conventional LLM fine-tuning, on the MixATIS and MixSNIPS datasets. ECLM demonstrated robust performance across various intent counts, particularly excelling in complex multi-intent scenarios.


\section{Acknowledgements}
This work was supported by the National Natural Science Foundation of China (62306119 and 71472068) and the Science and Technology Projects in Guangzhou (2025A04J3436).

\section*{Limitations}
(1) \textit{Scaling up Model Size of ECLM}: Due to computational resource constraints, we were unable to experiment with ECLM models larger than 8 billion parameters. However, we believe that scaling to larger model sizes could potentially yield further improvements in performance. Recent trends in language model research suggest that larger models often demonstrate enhanced capabilities across various NLP tasks. Future work with access to more substantial computational resources could explore the impact of increased model size on ECLM's performance in multi-intent SLU tasks.
(2) \textit{Prospects for Improvement through Data Curation and Prompt Optimization}: Our current research framework does not extend to the advanced strategies of selective data curation or intricate prompt engineering. Recognizing this as a limitation, we propose that future investigations will embrace these crucial techniques. 

\bibliography{anthology,custom}
\bibliographystyle{acl_natbib}

\newpage
\appendix

\section{Appendix}
\label{sec:appendix}

\subsection{Experiments on the TOPv2 Dataset}

To further assess generalization to semantic parsing benchmarks, we conducted experiments on the \textsc{TOPv2} dataset \citep{chen-etal-2020-low-resource}, focusing on the \textit{Alarm} and \textit{Weather} domains. We used 20k training samples and 2k evaluation samples per domain, following the same model configuration as in our main experiments.

\begin{table}[ht]
\centering
\small
\setlength{\tabcolsep}{4pt}
\begin{tabular}{llccc}
\toprule
\textbf{Domain} & \textbf{Model} & \textbf{Slot F1} & \textbf{Intent Acc} & \textbf{Overall Acc} \\
\midrule
\multirow{2}{*}{Alarm} 
  & ECLM         & \textbf{0.87} & \textbf{0.95} & \textbf{0.87} \\
  & Vanilla SFT  & 0.84 & 0.93 & 0.83 \\
\midrule
\multirow{2}{*}{Weather} 
  & ECLM         & \textbf{0.97} & \textbf{0.96} & \textbf{0.92} \\
  & Vanilla SFT  & 0.55 & 0.95 & 0.60 \\
\bottomrule
\end{tabular}
\caption{Performance comparison on the TOPv2 dataset.}

\label{tab:topv2}
\end{table}

As shown in Table~\ref{tab:topv2}, our method significantly outperforms the Vanilla SFT baseline in terms of overall accuracy. The improvement in the Weather domain is particularly notable, likely due to the increased complexity and diversity of slot annotations in weather-related utterances.


\subsection{Additional Case Illustrations}
\label{Case}




\begin{figure*}[t]
    \centering
    \includegraphics[width=0.85\linewidth]{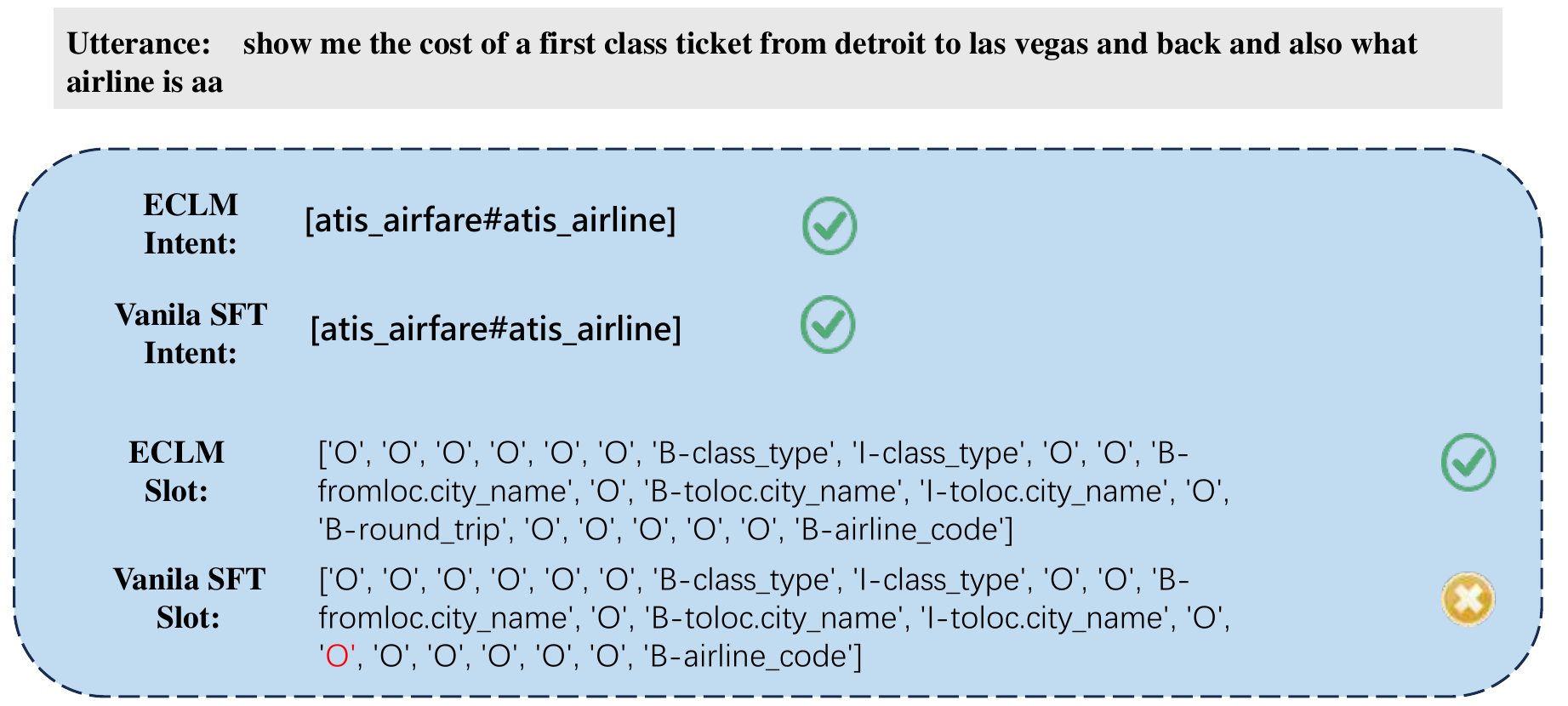}
    \caption{Case 1.}
    \label{fig:case1}
    
    \vspace{0.5em}
    
    \includegraphics[width=0.85\linewidth]{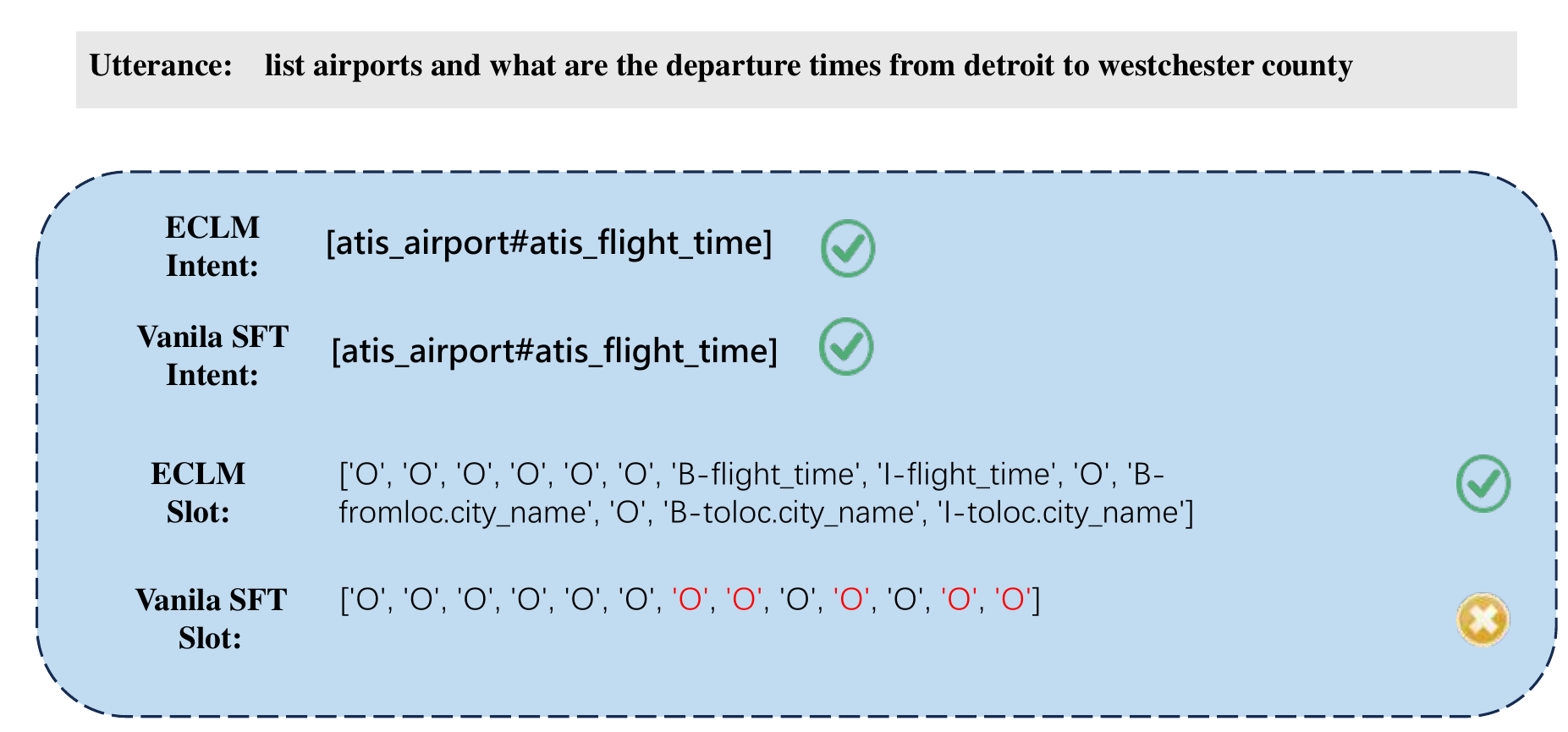}
    \caption{Case 2.}
    \label{fig:case2}
    
    \vspace{0.5em}
    
    \includegraphics[width=0.85\linewidth]{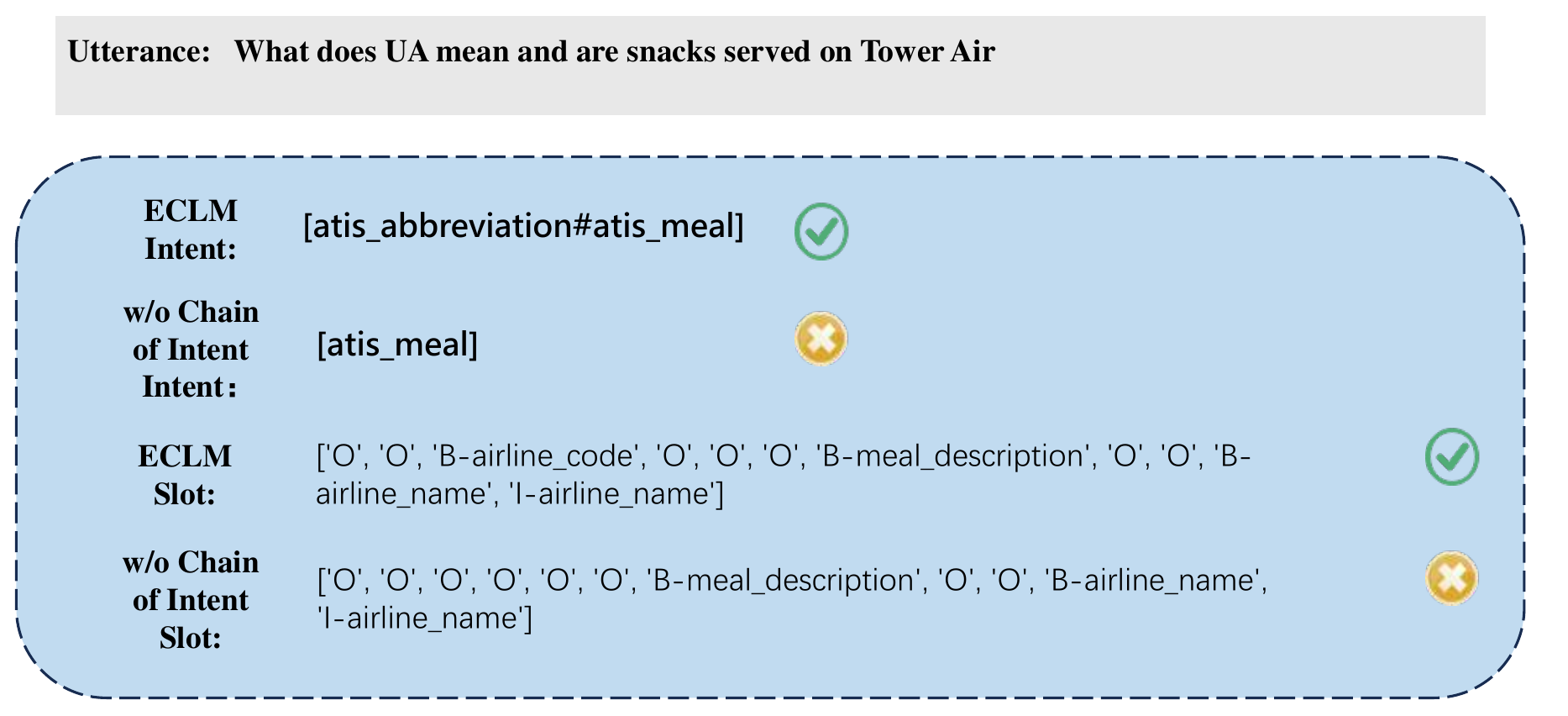}
    \caption{Case 3.}
    \label{fig:case3}
\end{figure*}

\end{document}